\UseRawInputEncoding

\documentclass[10pt, conference, compsocconf]{IEEEtran}
\usepackage{svg}
\usepackage{float}
\usepackage{booktabs}
\usepackage{subcaption}
\usepackage{amssymb}
\usepackage{hyperref}
\usepackage{amsmath}
\usepackage{pifont}
\usepackage[linesnumbered,ruled,vlined]{algorithm2e}
\usepackage{subcaption}
\usepackage{graphicx}
\usepackage{cite}

\ifCLASSINFOpdf
\else
\fi

\usepackage{tikz}
\usepackage{tabularx}
\usepackage{multirow}

\usepackage{xcolor}

\begin{document}
%
\title{Domain-Guided Masked Autoencoders for Unique Player Identification}



%
\author{\IEEEauthorblockN{Bavesh Balaji\IEEEauthorrefmark{1}\IEEEauthorrefmark{2},
Jerrin Bright\IEEEauthorrefmark{1},
Sirisha Rambhatla\IEEEauthorrefmark{2},
Yuhao Chen\IEEEauthorrefmark{1},
Alexander Wong\IEEEauthorrefmark{1},
John Zelek\IEEEauthorrefmark{1} and
David A Clausi\IEEEauthorrefmark{1}}
\IEEEauthorblockA{\IEEEauthorrefmark{1} Vision and Image Processing Lab, University of Waterloo}
\IEEEauthorblockA{\IEEEauthorrefmark{2} Critical ML Lab, University of Waterloo}
{\{bbalaji, jerrin.bright, sirisha.rambhatla, yuhao.chen1, alexander.wong, jzelek, dclausi\}}@uwaterloo.ca

}


\maketitle

\begin{abstract}
Unique player identification is a fundamental module in vision-driven sports analytics. Identifying players from broadcast videos can aid with various downstream tasks such as player assessment, in-game analysis, and broadcast production. However, automatic detection of jersey numbers using deep features is challenging primarily due to: a) motion blur, b) low resolution video feed, and c) occlusions. With their recent success in various vision tasks, masked autoencoders (MAEs) have emerged as a superior alternative to conventional feature extractors. However, most MAEs simply zero-out image patches either randomly or focus on where to mask rather than how to mask. Motivated by human vision, we devise a novel domain-guided masking policy for MAEs termed \textit{d}-MAE to facilitate robust feature extraction in the presence of motion blur for player identification. We further introduce a new spatio-temporal network leveraging our novel \textit{d}-MAE for unique player identification. We conduct experiments on three large-scale sports datasets, including a curated baseball dataset, the SoccerNet dataset, and an in-house ice hockey dataset. We preprocess the datasets using an upgraded keyframe identification (KfID) module by focusing on frames containing jersey numbers. Additionally, we propose a keyframe-fusion technique to augment keyframes, preserving spatial and temporal context. Our spatio-temporal network showcases significant improvements, surpassing the current state-of-the-art by 8.58\%, 4.29\%, and 1.20\% in the test set accuracies, respectively. Rigorous ablations highlight the effectiveness of our domain-guided masking approach and the refined KfID module, resulting in performance enhancements of 1.48\% and 1.84\% respectively, compared to original architectures.
\end{abstract}


%
\IEEEpeerreviewmaketitle

\section{Introduction}

\begin{figure}[t]
  \centering
  \begin{tikzpicture}
    \node at (0,0) {\includegraphics[width=\linewidth]{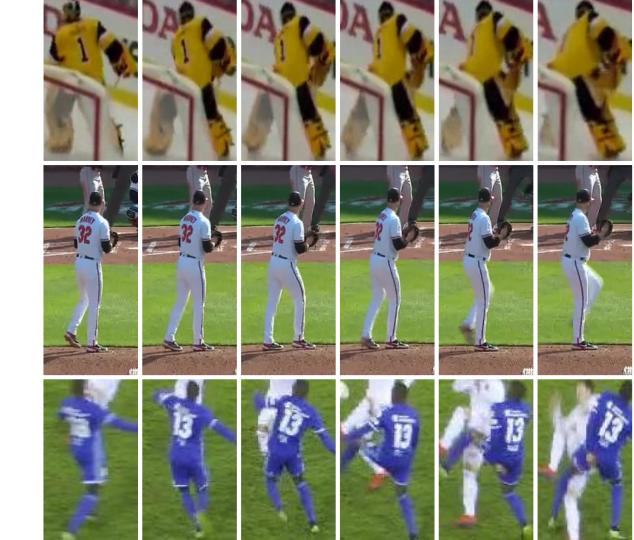}};
    \node[font=\small, rotate=90] at (-4,2.5) {Ice Hockey};
    \node[font=\small, rotate=90] at (-4,0) {Baseball};
    \node[font=\small, rotate=90] at (-4,-2.5) {Soccer};
  \end{tikzpicture}
   \caption{Example frames of various tracklets from three large-scale sports datasets showcasing the challenges: motion blur, occlusions and low-resolution.}
   \label{fig:dataset}
\end{figure}
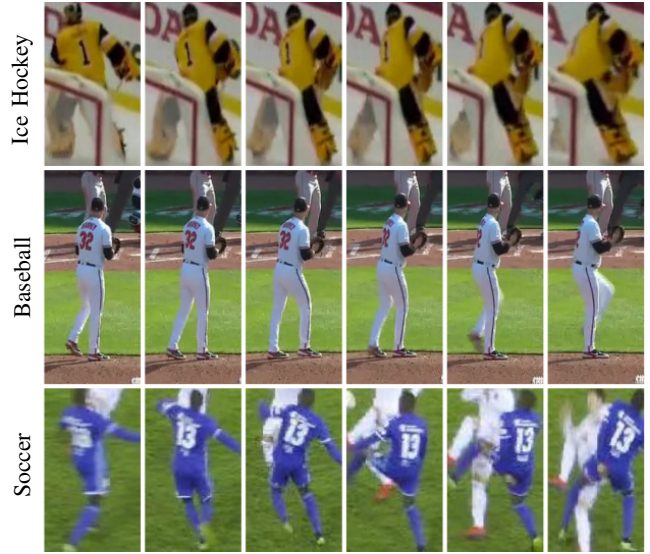

Unique player identification in real-world broadcast videos is a critical challenge that has been extensively researched over the years \cite{vats-trans, mmjersey, Liu2019PoseGuidedRF, STN, liu2022deep}. Accurate identification of individual players in sports holds significant importance in various contexts ranging from performance analysis to tactical evaluation of games \cite{vats-track, Lu} by coaches and scouts. However, precise player identification is inherently challenging due to the fast-paced nature of most sports. The rapid movements and unpredictable actions of the players often result in motion blur and occlusions, as shown in Fig. \ref{fig:dataset}. Addressing these challenges necessitate the development of adaptive techniques capable of effectively handling motion blur, occlusions, and other common issues encountered in real-world sporting environments.

\begin{figure*}
  \centering

  \begin{tikzpicture}
    \node at (0,0) {\includegraphics[width=\linewidth]{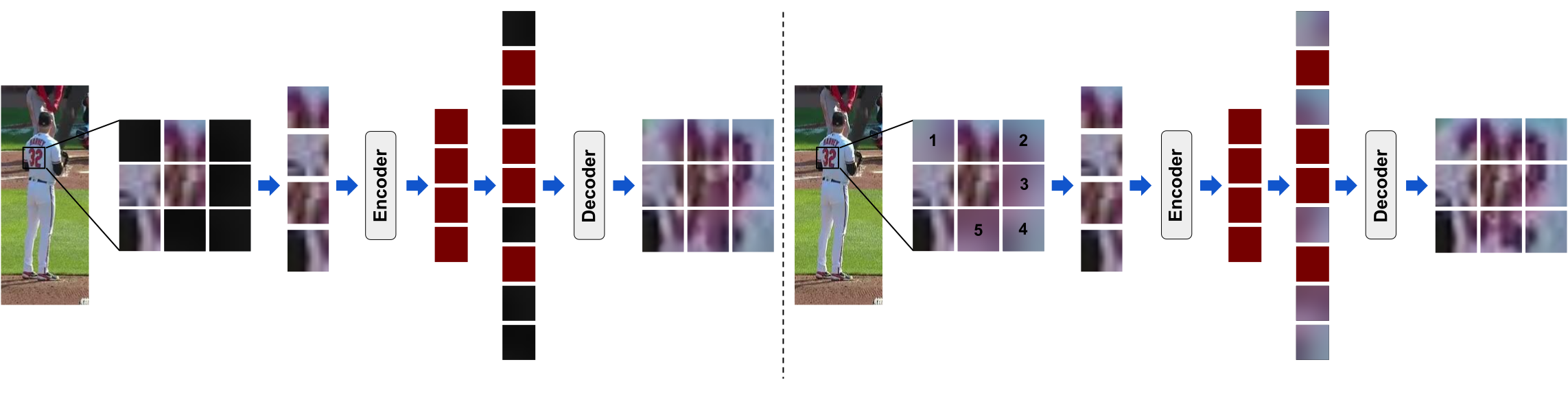}};
    \node[font=\normalsize] at (-5.2,-1.7) {(a) Existing MAEs};
    \node[font=\normalsize] at (4.1,-1.7) {(b) \textit{d}-MAE (Ours)};
  \end{tikzpicture}
  
   \vspace{-0.5em}
   \caption{\textbf{Comparison of \textit{d}-MAE with existing MAEs}. (a) Existing MAEs zero-out/ blackout patches randomly while (b) We introduce motion blur artifacts on random patches. The masked patches in (b) are numbered from 1-5.}
   \label{fig:teaser}
\end{figure*}

Previous works on jersey number recognition including \cite{STN, vats2021multitask, Liu2019PoseGuidedRF} have primarily focused on extracting spatial features from static images. However, these approaches often rely on conventional convolutional networks \cite{resnet101}, which struggle to handle the aforementioned non-idealities. Recent works including \cite{vats-trans, chan}, leverage the temporal features using Transformers \cite{attention} and LSTMs \cite{Hochreiter1997LongSM} to identify players from video. Despite these advancements, the absence of jersey numbers in numerous frames has been identified to be detrimental to existing models \cite{mmjersey}. To tackle this issue, \cite{mmjersey} proposed a unique keyframe identification (KfID) module to effectively capture keyframes from player tracklets. While KfID effectively isolates frames with clear jersey numbers, its strict thresholding criteria and sole usage of color for frame association leads to the loss of desirable frames, thereby reducing the amount of data.
Nevertheless, within the realm of unique player identification,  no prior work, to the best of our knowledge, has specifically focused on effectively capturing spatial context from sports video that deals with common problems like motion blur and occlusion. 


Recently, masked autoencoders (MAEs), inspired from masked language modeling \cite{GPT, BERT}, have emerged as a promising self-supervised learning method for robust spatial feature extraction utilizing vision transformers \cite{Dosovitskiy2020AnII}. MAEs \cite{He2021MaskedAA} aim to learn semantic representations by zeroing-out random patches and reconstructing the input image using the visible patches. Although MAEs are robust to occlusions, they perform suboptimally in the presence of motion blur as illustrated in Fig. \ref{fig:teaser}(a). This raises critical questions concerning the design of MAEs in learning representations from different \textit{domains} such as sports. \textit{Where should we mask?} \textit{How should we mask?} \textit{Is zeroing-out the only way to do it?} While significant research has tackled the first question \cite{AdaMAE, attmae, AutoMAE}, the latter two remain underexplored. 

To address these underexplored questions for jersey number recognition, we propose a novel spatio-temporal network that utilizes MAEs with a new masking strategy specific to the domain of sports (domain-guided). To improve our MAE's robustness to motion blur, we introduce motion blur artifacts on random patches instead of zeroing them out. This masking policy proves to be advantageous in extracting effective visual representations, considering the prevalence of motion blur in sports data. Fig. \ref{fig:teaser} illustrates the different masking strategies employed by the vanilla-MAE and our proposed approach. Doing so, we outperform the state-of-the-art jersey number recognition networks by 8.58\%, 4.29\% and 1.2\% on three large-scale sports datasets. Furthermore, we quantitatively validate our masking strategy against conventional MAEs, demonstrating superior performance. Moreover, we enhance the KfID module proposed in \cite{mmjersey} to capture keyframes containing vital information, resulting in a significant improvement of 1.84\% compared to the existing module. In summary our contributions include the following: 


\begin{enumerate}
    \item We introduce a novel jersey number recognition network that utilizes MAEs coupled with a transformer decoder to capture robust features from low-resolution blurred tracklets.
    \item We propose a new domain-guided masking strategy, termed \textit{d}-MAE, specifically tailored to player identification, enhancing model robustness to motion blur. 
    \item We refine the KfID module by improving its jersey number localization and its ability to capture fine-grained semantic representations of keyframes. 
    \item Addressing the issue of limited data, we introduce a keyframe fusion technique to augment meaningful data, thereby enriching the training process.
    \item We validate that our model outperforms existing state-of-the-art methods on three large-scale datasets spanning different sports.
\end{enumerate}

\section{Related Work} \label{works}

\subsection{Masked Image Modeling (MIM)}
Early classical approaches such as classical inpainting \cite{Bertalmo2000ImageI, Osher2005AnIR} and texture synthesis \cite{Barnes2009PatchMatchAR, Efros1999TextureSB} mainly focused on denoising small portions of an image. Hence, these were not very effective in reconstructing very large regions of mask. They also had challenges with filling in objects that were partially occluded. Therefore, to address these issues, Vincent \textit{et al.} \cite{Vincent2010StackedDA} use denoising autoencoders to reconstruct corrupted images. \textit{Pathak et al.} introduced context encoders to fill out large holes(rectangular masks) created in an image. Following the success of Masked language modeling using transformers in NLP, a similar approach to predicting missing pixels was then pursued in computer vision. Chen et al. \cite{Chen2020GenerativePF} use a sequence transformer to auto-regressively predict pixels. Other works such as \cite{He2021MaskedAA, Bao2021BEiTBP, Dosovitskiy2020AnII} focus on representing an image as discrete tokens and mask them randomly to reconstruct them. However, all of the above methods focus on one masking strategy:- zeroing-out random image patches. More recent works such as \cite{attmae} explore the use of attention maps, while \cite{AutoMAE} uses an adversarial mask-generator to learn where to mask. Bandara \textit{et al.} \cite{AdaMAE} uses a novel token sampling strategy to sample tokens with high spatiotemporal information. 

\textbf{Research Gap in MIM.} The above methods on masked autoencoders are focused on the \textit{where to mask} sampling strategy rather than the \textit{how to mask} strategy. We explore into this unexplored realm by developing a domain-guided masking strategy specifically designed to address the challenge of highly prevalent motion blur in sports videos. 

\subsection{Jersey Number recognition from static images}

The advent of deep learning facilitated the use of jersey numbers rather than player appearances for unique player identification. Earlier works \cite{Gerke2015SoccerJN, STN, Liu2019PoseGuidedRF} use CNNs to predict the jersey number: Gerke \textit{et al.} \cite{Gerke2015SoccerJN} recognize jersey numbers directly from soccer images while Li \textit{et al.} \cite{STN} and Liu \textit{et al.} \cite{Liu2019PoseGuidedRF} propose a unified solution to detect and classify jersey numbers using Spatial Transformer Networks (STNs) and pose-guided recurrent CNNs, respectively. Vats \textit{et al.} \cite{vats2021multitask} leverage a multi-task loss function to perform holistic and digit-wise predictions. Bhargavi \textit{et al.} \cite{bhargavi2022knock} present a multi-stage network that takes advantage of pose to localize jersey numbers before detecting them using a secondary classifier.

\subsection{Jersey Number recognition from player tracklets}

More recent works \cite{vats-trans, liu2022deep, chan} aim at leveraging the temporal aspect and capture temporal cues from tracklets. Vats \textit{et al.} \cite{vats-trans} developed a transformer-based architecture to recognize jersey numbers from Ice Hockey videos.  Chan \textit{et al.} \cite{chan}, on the other hand, utilize LSTMs to extract temporal characteristics from player tracklets. Furthermore, they also employ 1D CNNs as a late score-level fusion method for classification. Liu \textit{et al.} \cite{liu2022deep} adopt a two-stage approach to perform the detection and classification of players from American football videos. Balaji \textit{et al.} \cite{mmjersey} propose a novel KfID module to remove redundant frames that contain no essential information about the jersey number.

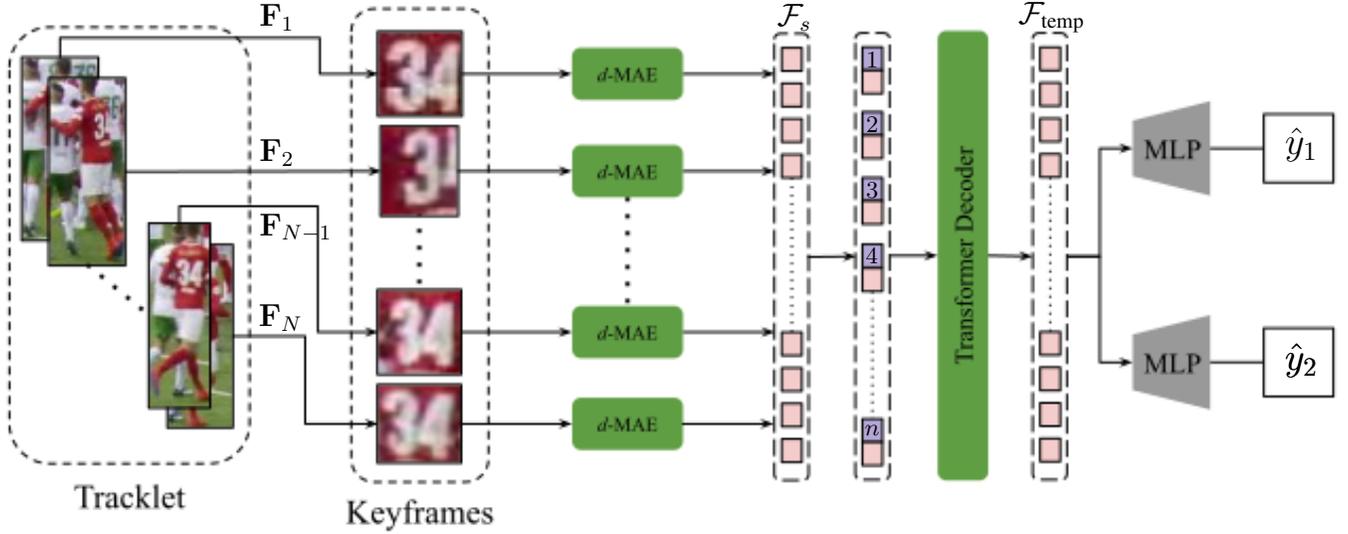
\begin{figure*}
  \centering

  \begin{tikzpicture}
    \node at (0,0) {\includegraphics[width=\linewidth]{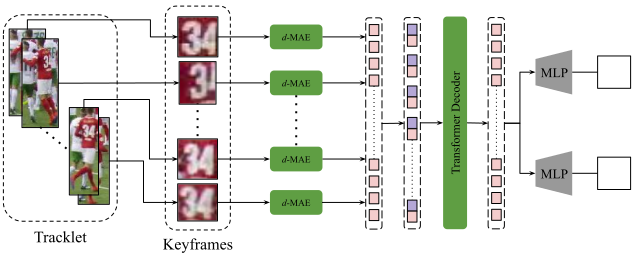}};
    \node[font=\small] at (2.665,2.93) {$1$};
    \node[font=\small] at (2.665,2.07) {$2$};
    \node[font=\small] at (2.665,1.20) {$3$};
    \node[font=\small] at (2.66,0.32) {$4$};
    \node[font=\small] at (2.665,-2) {$n$};
    \node[font=\Large] at (8.37,1.8) {$\hat{y}_1$};
    \node[font=\Large] at (8.37,-1.07) {$\hat{y}_2$};
    \node[font=\Large] at (8.37,-1.07) {$\hat{y}_2$};
    \node[font=\large] at (1.63,3.5) {$\mathcal{F}_s$};
    \node[font=\large] at (5.05,3.5) {$\mathcal{F}_{\textrm{temp}}$};
    \node[font=\large] at (-5.25,3.5) {$\mathbf{F}_1$};
    \node[font=\large] at (-5.25,1.7) {$\mathbf{F}_2$};
    \node[font=\large] at (-5.,0.7) {$\mathbf{F}_{N-1}$};
    \node[font=\large] at (-5.2,-0.5) {$\mathbf{F}_{N}$};
  \end{tikzpicture}
  
   \vspace{-1em}
   \caption{\textbf{Overall architecture}. Given a tracklet $\mathbb{T}$ consisting of $N$ frames, we pass $\mathbb{T}$ through the KfID module to extract $n \leq N$ keyframes that contain the jersey number. Each keyframe is passed as an input to our \textit{d}-MAE encoder to extract spatial features $\mathcal{F}_s$. These features are then fed to the temporal transformer decoder to extract temporal features $\mathcal{F}_{\textrm{temp}}$. Two classification heads are utilized to compute the predicted digits of the jersey number $\hat{y}_1$ and $\hat{y}_2$ respectively.} 
   \label{fig:overview}
\end{figure*}

\textbf{Research Gap.} While previous methods on player identification from tracklets prioritize enhancing the temporal representation of their networks for accurate jersey number recognition, our focus lies on improving spatial feature extraction, which contains richer information and enhances our temporal decoder. To achieve this, we leverage MAEs tailored for sports data. The proposed system depicted in Fig. \ref{fig:overview} is designed to handle motion blur and occlusions, thereby aiding in determining player identities in a more reliable manner.

\section{Methodology} \label{method}
The overview of the proposed architecture is shown in Fig. \ref{fig:overview}. Initially, the frames of a player tracklet are processed in the KfID module, where keyframes crucial for identifying jersey numbers are extracted. Subsequently, these keyframes are passed to our \textit{d}-MAE encoder, which captures features with rich semantic representations of each keyframe. These spatial features are then passed to the temporal transformer decoder, which extracts temporal cues and predicts the jersey number associated with the tracklet. More details on the individual modules are explained in this section.

\subsection{Domain-guided Masked Autoencoders}


The application of MAE becomes particularly significant for jersey number detection due to the dynamic nature of the game. Players are frequently in motion leading to challenges such as occlusion, low-visibility and motion blur. However, we recognize that the conventional masking policy (zeroing-out) used in MAEs do not encompass the diverse conditions encountered in real-world scenarios. For instance, viewing an image through a window filled with water droplets provides a form of masking that is not similar to zeroing-out image patches. Similarly, broadcast feeds of fast-paced sports like Soccer, Basketball, or Ice Hockey highlight the occurrence of such visual distortions and blurring effects that affect visual clarity. 

Motivated by the need to recover missing or occluded spatial information to acknowledge these diverse scenarios, we build on the proficiency of MAE to reconstruct missing patches within the pixel space by introducing a domain-guided masking strategy. Particularly, during the pre-training stage, we incorporate motion blur to the patches instead of simply zeroing-out them, thereby infusing domain knowledge in the process. This approach facilitates reliable and accurate prediction of jersey numbers in dynamic sports scenarios. We incorporate an additional supervision to the pre-training objective, improving the feature extraction process. 




\textbf{Pre-training.} The input image $\mathbf{I} \in \mathbb{R}^{H \times W \times C}$ is split into $K$ patches $\mathbb{I} = \{\mathbf{I}_k \in \mathbb{R}^{P^2 \times D}\}_{k=1}^K$ where $P$ is the patch size, $D$ is the embedding dimension and $K = HW / P^2$. A random subset of the patches $\mathbb{S} \subseteq \mathbb{I}$ are then masked by introducing motion blur artifacts to the pixels, resulting in the set
\begin{equation*}
    \label{eq:1}
    \mathbb{I}_{\textrm{masked}} = 
    \{m(\mathbf{x}):\mathbf{x} \in \mathbb{S}\},
\end{equation*}
where $m:\mathbb{R}^{P^2 \times C} \rightarrow \mathbb{R}^{P^2 \times C}$ is the mask applied to random patches. The unmasked patches $\mathbb{I}_{\textrm{unmasked}} = \mathbb{I} \setminus \mathbb{S}$ are then converted to unmasked tokens
\begin{equation*}
    \label{eq:2}
    \text{{\fontfamily{qcr}\selectfont
        [unmask]
    }} = \textrm{concat}(\mathbf{x}:\mathbf{x} \in \mathbb{I}_{\textrm{unmasked}})
\end{equation*} 
and passed to the MAE encoder to extract latent spatial features \begin{equation*}
    \label{eq:3}
    \mathcal{F}_s = f(\text{{\fontfamily{qcr}\selectfont
        [unmask]}}),  
\end{equation*} where $f:\mathbb{R}^{K \times P^2 \times D} \rightarrow \mathbb{R}^{K \times P^2 \times D}$ denotes the MAE encoder. The masked tokens \begin{equation*}
    \label{eq:4}
    \text{{\fontfamily{qcr}\selectfont
        [mask]}} = \textrm{concat}(\mathbf{x}:\mathbf{x} \in \mathbb{I}_{\textrm{masked}})
\end{equation*} are then used along with the latent features to generate the reconstructed image \begin{equation*}
    \label{eq:5}
    \mathbf{\hat{I}} = g(\textrm{concat}(\text{{\fontfamily{qcr}\selectfont
        [mask]}}, \mathcal{F}_s)),
\end{equation*}
where $g:\mathbb{R}^{K \times P^2 \times D} \rightarrow \mathbb{R}^{H \times W \times C}$ denotes the MAE decoder. 




To induce motion blur in the selected patches, we employ an oriented motion blur filter $\mathbf{K}'$ characterized by two parameters: the angle of rotation ($\omega$) and the scale factor ($s_f$). The filter is centered at $(k_s / 2, k_s / 2)$ where $k_s$ is the kernel size. Eq. \eqref{eq:mblur} denotes the motion blur filter used to apply motion blur on image patches. 
    
\begin{equation}
    \label{eq:mblur}
    m(\mathbf{I_k})(x, y) = \dfrac{\mathbf{K}'}{{\sum}_{i,j} \mathbf{K}'_{ij}} * \mathbf{I}_k(x,y)
\end{equation}

\begin{equation}
    \label{eq:rot}
    \mathbf{R} = \left[ \begin{smallmatrix}
    s_f\cos(\omega)  & - s_f\sin(\omega) & \tfrac{k_s}{2} (1 - s_f\cos(\omega)) + \tfrac{k_s}{2} s_f\sin(\omega)  \\
    s_f\sin(\omega)  & s_f\cos(\omega)  & \tfrac{k_s}{2}(1 - s_f\cos(\omega)) - \tfrac{k_s}{2} s_f \cos(\omega) 
    \end{smallmatrix} \right]
\end{equation}

where $*$ denotes the convolution operation and $m(.)$ is the masking strategy we employ at every pixel position $(x, y)$ of an image patch $\mathbf{I}_k$. The definition of the rotation matrix $\mathbf{R}$ used to generate the oriented filter $\mathbf{K}'$ is shown in \eqref{eq:rot}.

This tailored approach facilitates our \textit{d}-MAE in capturing crucial cues necessary for the accurate reconstruction of the keyframes in the presence of motion blur. 


\subsection{Transformer decoder}

To capture the temporal cues within the tracklet, we extend our MAE module, by introducing a transformer decoder. Specifically, after the pre-training stage, we discard the decoder of \textit{d}-MAE during finetuning, and pass the original unmasked keyframes directly to the \textit{d}-MAE encoder. The extracted latent spatial features $\mathcal{F}_s$ are fed to the temporal transformer decoder to perform jersey number recognition. Leveraging the standard vision transformer (ViT) architecture for our decoder, we utilize self-attention to capture long-range dependencies between the spatial features of different frames within a player tracklet. By employing the self-attention mechanism on the spatial tokens, we facilitate the model's ability to understand the global context and intricate connections between keyframes of a tracklet. The resulting representation $\mathcal{F}_{\textrm{temp}}$, encapsulates rich cues on the jersey number, which are crucial for player identification.

\subsection{Keyframe Identification}

The Keyframe Identification (KfID) module was proposed in \cite{mmjersey} with the objective to capture critical frames of a player tracklet where jersey numbers are visible. It consists of three critical components -- 1) \textit{Jersey Number Localization (JNL)} localizes all digits in a particular frame $\mathbf{F}_i$; (2) \textit{RoI-based filtering} captures digits of our player of interest by filtering all the digits that are within a preset region of interest (RoI) (3) \textit{Local Histogram Correlation (LHC)} creates the holistic representation of the jersey number of our player of interest by merging digits detected in a frame $\mathbf{F}_i$; and (4) \textit{Global Histogram Correlation (GHC)} clusters different frames using their spatial color (HSV) layout to find keyframes that contain the jersey number of our player of interest. 

In mathematical terms, given a player tracklet\\ 
$\mathbb{T} = \{\mathbf{F}_i : \mathbf{F}_i \in \mathbb{R}^{H \times W \times3}\}_{i=1}^N$ consisting of $N$ frames, 

\begin{equation}
    \text{KfID}(\mathbb{T})  = \mathbb{T} \setminus \{\mathbf{F}_{n_1}, \mathbf{F}_{n_2}, ..., \mathbf{F}_{n_k}\}
\end{equation}
where $\{\mathbf{F}_{n_1}, \mathbf{F}_{n_2}, .., \mathbf{F}_{n_k}\}$ denotes the set of noisy frames that need to be removed as they contain no relevant information regarding the jersey number and affect the performance of the models. 

While the KfID module contributes to a significant improvement over the existing frameworks for jersey number identification by filtering out undesirable frames from the tracklet, we observe that its JNL module fails to recognize smaller digits and provides false positives in noisy scenarios. Furthermore, the GHC module, relying solely on color for frame association, tends to produce spurious clusters. To address these issues and further enhance the KfID module, we introduce architectural modifications to the JNL and GHC modules. Additionally, we incorporate a keyframe fusion-based data augmentation to address challenges with limited labelled data.

\textbf{JNL module.} The vanilla KfID module utilizes a finetuned YOLOv5 \cite{yolov5} model to detect digits from the input frames of player tracklets. However, it tends to perform poorly in noisy and crowded scenarios, particularly when dealing with the detection of smaller digits. To address these issues, we use a pretrained deformable-DETR \cite{deformable-detr} model and finetune it on our digit detection dataset captured from SoccerNet images for reliable digit detection. We chose the deformable-DETR network because of its ability to accommodate high-resolution feature maps in a transformer network, enabling capturing of fine-grained representations of smaller objects. 

\textbf{GHC Module.} Balaji \textit{et al.} \cite{mmjersey} utilize the spatial color layout of the holistic detections from different frames of a tracklet and employ clustering techniques to associate frames containing the jersey number of the player of interest. However, when multiple players from the same team appear in a tracklet, this approach fails since it solely leverages color to cluster frames irrespective of their jersey numbers. To overcome this challenge, we propose extracting deep semantic features such as shape and texture of each frame using a ResNet backbone. These features are then clustered to accurately identify the keyframes within a player tracklet. 

\textbf{Keyframe Fusion.} To tackle the challenge of limited labeled data from the incorporation of the KfID module, we introduce a strategic \textit{fusion augmentation} technique to counter the shortcomings. Specifically, we implement this augmentation by randomly selecting $n$ consecutive frames from a specific sequence within a tracklet and merging them together. This fusion captures rich visual representation of the digits especially in scenarios with noise or fast-paced movements in the keyframes. By fusing all the frames within a sequence of a tracklet, we preserve the temporal flow of information. This ensures that the fusion augmentation doesn't introduce additional noise by fusing frames that are too distant to each other. 

\subsection{Loss Functions}

\textbf{Siamese Loss.} Inspired by the existing literature from 3D vision \cite{humannerf, gauhuman}, we employ an additional Siamese objective apart from the MSE loss to supervise the reconstructed image from the MAE decoder while pre-training it. Given a prediction ($\hat{\mathbf{I}}$) and groundtruth ($\mathbf{I}$) image, we extract features from both the images using a pretrained feature extractor $h(\cdot)$ and minimize the discrepancy using $\ell_1$-loss.  

\begin{equation}
    \mathcal{L}_{\textrm{siamese}} = ||h(\hat{\mathbf{I}}) - h(\mathbf{I})||_1
    \label{eq:lpips}
\end{equation}

We use ResNet as $h(.)$ to extract features from the images and incorporate $\ell_1$-loss instead of the MSE metric used in \cite{gauhuman}. Ablations on the different setups are detailed in Table \ref{tab:ploss}.
 
The total loss of our MAE network in pre-training stage is as follows:
\begin{equation}
    \mathcal{L}_{\textrm{mae}} = \sigma_1~\lvert\lvert\mathbf{\hat{I}} - \mathbf{I} \rvert\rvert_2 + \sigma_2~\mathcal{L}_{\textrm{siamese}} \label{eq:tot}
\end{equation}
where $\sigma_1$ and $\sigma_2$ are learnable weights. 

\textbf{Multi-head Classification Loss.} We employ a multi-task loss function using cross-entropy to effectively classify jersey numbers \cite{mmjersey} from the transformer decoder, as shown in \eqref{eq:mtl}, where $y_1 \in \mathbb{R}^{11}$ and $y_2 \in \mathbb{R}^{11}$ are ground-truth digits of the jersey number (10 digits + 1 null class), and $\Hat{y_1} \in \mathbb{R}^{11}$ and $\Hat{y_2} \in \mathbb{R}^{11}$ are the predictions made by the spatiotemporal network.

\begin{equation}
    \mathcal{L}_{\textrm{class}} = -\sum_{i=0}^{10} y_1^i\log\Hat{y}_1^i -\sum_{j=0}^{10} y_2^j\log\hat{y}_2^j \label{eq:mtl}
\end{equation}

\begin{table*}
  \caption{Dataset split-up for training, validation and testing}
  \label{tab:combined-datasets}
  \centering
  \begin{tabular}{lccc|ccc|ccc}
    \toprule
    & \multicolumn{3}{c}{\textbf{SoccerNet Dataset}} & \multicolumn{3}{c}{\textbf{Ice Hockey Dataset}} & \multicolumn{3}{c}{\textbf{Baseball Dataset}} \\
    \cmidrule(lr){2-4} \cmidrule(lr){5-7} \cmidrule(lr){8-10}
    \textbf{Dataset} & \textbf{Tracklets} & \textbf{Images} & \textbf{Keyframes} & \textbf{Tracklets} & \textbf{Images} & \textbf{Keyframes} & \textbf{Tracklets} & \textbf{Images} & \textbf{Keyframes} \\
    \midrule
    Train & 1,141 & 587,543 & 71,021 & 2829 & 540,339 & 162,101 & 105 & 21,050 & 18,344\\
    Validation & 286 & 146,886 & 19,609 & 176 & 33,616 & 11,084 & 15 & 2,962 & 2,571\\
    Test & 1,211 & 565,758 & 70,445 & 505 & 96,455 & 28,937 & 30 & 5,988 & 4,640\\
    Challenge & 1,426 & 750,092 & 101,307 & - & - & - & - & - & -\\
    \midrule 
    Total & 4,064 & 2,052,306 & 262,382 & 3510 & 670,410 & 202,122 & 150 & 30,000 & 25,555 \\
    \bottomrule
  \end{tabular}
\end{table*}

\begin{table*}
\caption{Quantitative Results}
  \begin{subtable}{0.6\linewidth}
    \centering
     \caption{Comparison of our model with existing state-of-the-art on the three datasets.}
    \begin{tabular}{lll|l|l}
      \toprule
      & \multicolumn{2}{c}{\textbf{SoccerNet}} & \multicolumn{1}{c}{\textbf{Ice Hockey}} & \multicolumn{1}{c}{\textbf{Baseball}} \\
      \cmidrule(lr){2-3} \cmidrule(lr){4-4} \cmidrule(lr){5-5}
      \textbf{Method} & \textbf{Test Acc} & \textbf{Challenge Acc} & \textbf{Test Acc} & \textbf{Test Acc}\\
      \midrule
      Gerke et al \cite{Gerke2015SoccerJN} & 32.57 & 35.79 & 61.20 & 64.47\\
      Vats et al \cite{vats2021multitask} & 46.73 & 49.88 & 83.17 & 87.61\\
      Li et al \cite{STN} & 47.85 & 50.60 & 81.15 & 88.29\\
      Vats et al \cite{vats-trans} & 52.91 & 58.45 & 85.14 & 89.46\\
      Balaji et al \cite{mmjersey}& 68.53 & 73.77 & 92.50 & 93.68\\
      \textbf{Ours} & \textbf{77.31} $\textcolor{green}{\mathbf{{\uparrow}8.58}}$ & \textbf{81.92} $\textcolor{green}{\mathbf{{\uparrow}8.15}}$& \textbf{96.79} $\textcolor{green}{\mathbf{{\uparrow}4.29}}$& \textbf{94.70} $\textcolor{green}{\mathbf{{\uparrow}1.02}}$\\
      \bottomrule
    \end{tabular}
   
    \label{tab:sota}
  \end{subtable}
  \hspace{30pt}
  \begin{subtable}{0.3\linewidth}
    \centering
     \caption{Impact of our MAE and masking strategy on the SoccerNet dataset}
      \begin{tabular}{lcc}
    \toprule
    \textbf{Backbone} & \textbf{Masking Strategy} & \textbf{Test Acc} \\
    \midrule
    ResNet-18 & - & 58.62 \\
    ResNet-34 & - & 61.29 \\
    ResNet-152 & - & 65.10\\
    MAE & Zeroing-Out & 75.83 \\
    MAE & Gaussian Blur & 76.47\\
    \textbf{MAE} & \textbf{Motion Blur} & \textbf{77.31}\\
    \bottomrule
  \end{tabular}
 
  \label{tab:heads}
  \end{subtable} 
\end{table*}

\section{Experiments} \label{exp}

\subsection{Datasets} \label{dataset}
We evaluate the performance of our model, along with the state-of-the-art methods, on three large-scale player tracklet datasets comprising videos from sports including Ice Hockey, Baseball, and Soccer. The dataset splitup is outlined in Table \ref{tab:combined-datasets}. Example tracklets for these datasets is illustrated in Fig. \ref{fig:dataset}.

\textbf{SoccerNet.} The SoccerNet player recognition dataset is the largest open source jersey number dataset in the world, comprising a total of 4,064 player tracklets. Each tracklet is predominantly focused on one player, and the label for the entire tracklet is the jersey number of that particular player. To facilitate model evaluation and training, the dataset has been partitioned into four distinct subsets: training, validation, testing, and challenge sets. The test and challenge sets are two different test sets sampled from different distributions of games. This helps us in evaluating the generalizability of models and their robustness to slight distribution shifts. 

\textbf{Baseball Dataset.} We have curated a player identification dataset from baseball videos, which is built based on the baseball 3D pose dataset introduced in \cite{mmblur}. The dataset comprises of 150 player tracklets sampled from over 1000 videos. Here, we utilize the videos from the aforementioned dataset and assign jersey number labels to each tracklet. 

\textbf{Ice Hockey Dataset.} We utilize the dataset introduced in \cite{vats-trans} to evaluate our model's performance on a fast-paced game with high motion blur and heavy equipment. The dataset consists of 3510 player tracklets generated from 84 NHL videos. The average length of a player tracklet is 191 frames, sampled at 30 fps. 

\subsection{Implementation Details}

\textbf{Data Augmentation.} We follow the data augmentation pipelines outlined in \cite{mmjersey} for the SoccerNet dataset and \cite{vats-trans} for the Ice Hockey dataset. For the Baseball dataset, our main augmentation strategies include color jitter and random rotation within $\pm$ 10 degrees. Subsequently, all images are resized to 224 $\times$ 224 and normalized using the ImageNet mean and standard deviation. 

\textbf{Model Settings.} We use the ViT-B variant of MAE for spatial feature extraction. Due to computational constraints, instead of pre-training it from scratch, we utilize a pretrained version of MAE that leverages the zeroing-out policy and further pre-train it on an in-house static jersey number dataset using our masking strategy. We follow the training pipeline outlined in the paper \cite{He2021MaskedAA} with modifications to incorporate the masking strategy to the MAE. The parameters $\omega$ and $s_f$ of the motion blur filter are empirically selected upon experimentation from a range of $0^{\circ} - 90^{\circ}$ and $0-2.0$ respectively. For the temporal transformer decoder, we use the standard ViT \cite{Dosovitskiy2020AnII} with 8 attention heads and 4 transformer layers for ideal performance. 
  
\textbf{Training Details.} We trained our model for 20,000 iterations on the SoccerNet and Ice Hockey dataset, and 10,000 iterations on the Baseball dataset with a batch size of 16. The AdamW optimizer was employed with an initial learning rate of 3e-4, with a learning rate scheduler that decreased the learning rate after every 2000 iterations for the initial 6000 iterations. All experiments were carried out using a single NVIDIA 2080Ti GPU with 12GB RAM. 

\subsection{Results} 

\textbf{Comparison with SOTA.} We conduct extensive experiments on the aforementioned datasets to assess the efficacy of our proposed architecture compared to existing state-of-the-art models on jersey number recognition. Table \ref{tab:sota} outlines our model's performance in comparison with existing works. The data processed by the KfID module serves as input to all the existing jersey number recognition networks for fair evaluation. Table \ref{tab:sota} demonstrates that our model consistently outperforms the existing techniques on all the three datasets. This demonstrates our model's ability to capture clearer spatial features, in dynamic sports scenarios characterized by challenges such as low-resolution, motion blur and occlusion. 

\begin{figure*}[t]
  \centering

  \begin{tikzpicture}
    \node at (0,0) {\includegraphics[width=\linewidth]{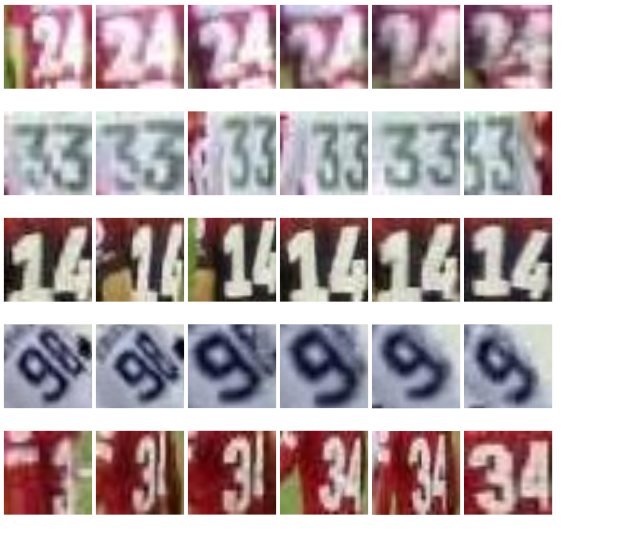}};
    \node[font=\normalsize] at (-4.9,1.85) {Prediction: 33};
    \node[font=\normalsize] at (-2.25,1.85) {Prediction: 33};
    \node[font=\normalsize] at (0.35,1.85) {Prediction: 33};
    \node[font=\normalsize] at (3,1.85) {Prediction: 33};
    \node[font=\normalsize] at (5.6,1.85) {Prediction: 33};
    \node[font=\normalsize] at (-7.5,1.85) {Prediction: 33};
    \node[font=\normalsize] at (-7.5,4.9) {Prediction: 24};
    \node[font=\normalsize] at (-4.9,4.9) {Prediction: 24};
    \node[font=\normalsize] at (-2.25,4.9) {Prediction: 24};
    \node[font=\normalsize] at (0.35,4.9) {Prediction: 24};
    \node[font=\normalsize] at (3,4.9) {Prediction: 24};
    \node[font=\normalsize] at (5.6,4.9) {Prediction: 24};
    \node[font=\normalsize] at (-7.5,-1.2) {Prediction: 14};
    \node[font=\normalsize] at (-4.9,-1.2) {Prediction: 16};
    \node[font=\normalsize] at (-2.25,-1.2) {Prediction: 14};
    \node[font=\normalsize] at (0.35,-1.2) {Prediction: 14};
    \node[font=\normalsize] at (3,-1.2) {Prediction: 14};
    \node[font=\normalsize] at (5.6,-1.2) {Prediction: 14};
    \node[font=\normalsize] at (-7.5,-4.25) {Prediction: 98};
    \node[font=\normalsize] at (-4.9,-4.25) {Prediction: 98};
    \node[font=\normalsize] at (-2.25,-4.25) {Prediction: 9};
    \node[font=\normalsize] at (0.35,-4.25) {Prediction: 9};
    \node[font=\normalsize] at (3,-4.25) {Prediction: 9};
    \node[font=\normalsize] at (5.6,-4.25) {Prediction: 9};
    \node[font=\normalsize] at (-7.5,-7.25) {Prediction: 3};
    \node[font=\normalsize] at (-4.9,-7.25) {Prediction: 31};
    \node[font=\normalsize] at (-2.25,-7.25) {Prediction: 31};
    \node[font=\normalsize] at (0.35,-7.25) {Prediction: 34};
    \node[font=\normalsize] at (3,-7.25) {Prediction: 34};
    \node[font=\normalsize] at (5.6,-7.25) {Prediction: 34};
    \node[font=\normalsize] at (7.7, -2.6) {\textbf{Pred}: 09};
    \node[font=\normalsize] at (7.7, -3.1) {\textbf{GT}: 98};
    \node[font=\normalsize] at (7.7, -5.6) {\textbf{Pred}: 34};
    \node[font=\normalsize] at (7.7, -6.1) {\textbf{GT}: 34};
    \node[font=\normalsize] at (7.7, 0.4) {\textbf{Pred}: 14};
    \node[font=\normalsize] at (7.7, -0.1) {\textbf{GT}: 14};
    \node[font=\normalsize] at (7.7, 3.4) {\textbf{Pred}: 33};
    \node[font=\normalsize] at (7.7, 2.9) {\textbf{GT}: 33};
    \node[font=\normalsize] at (7.7, 6.4) {\textbf{Pred}: 24};
    \node[font=\normalsize] at (7.7, 5.9) {\textbf{GT}: 24};
  \end{tikzpicture}
  
   \vspace{-1em}
   \caption{\textbf{Qualitative results}. Performance of our model on five different player tracklets from all the three datasets. We find our model's prediction for each image separately and for the entire tracklet (\textbf{Pred}). \textbf{GT} represents the ground-truth value for the entire tracklet.}
   \label{fig:qual}
\end{figure*}
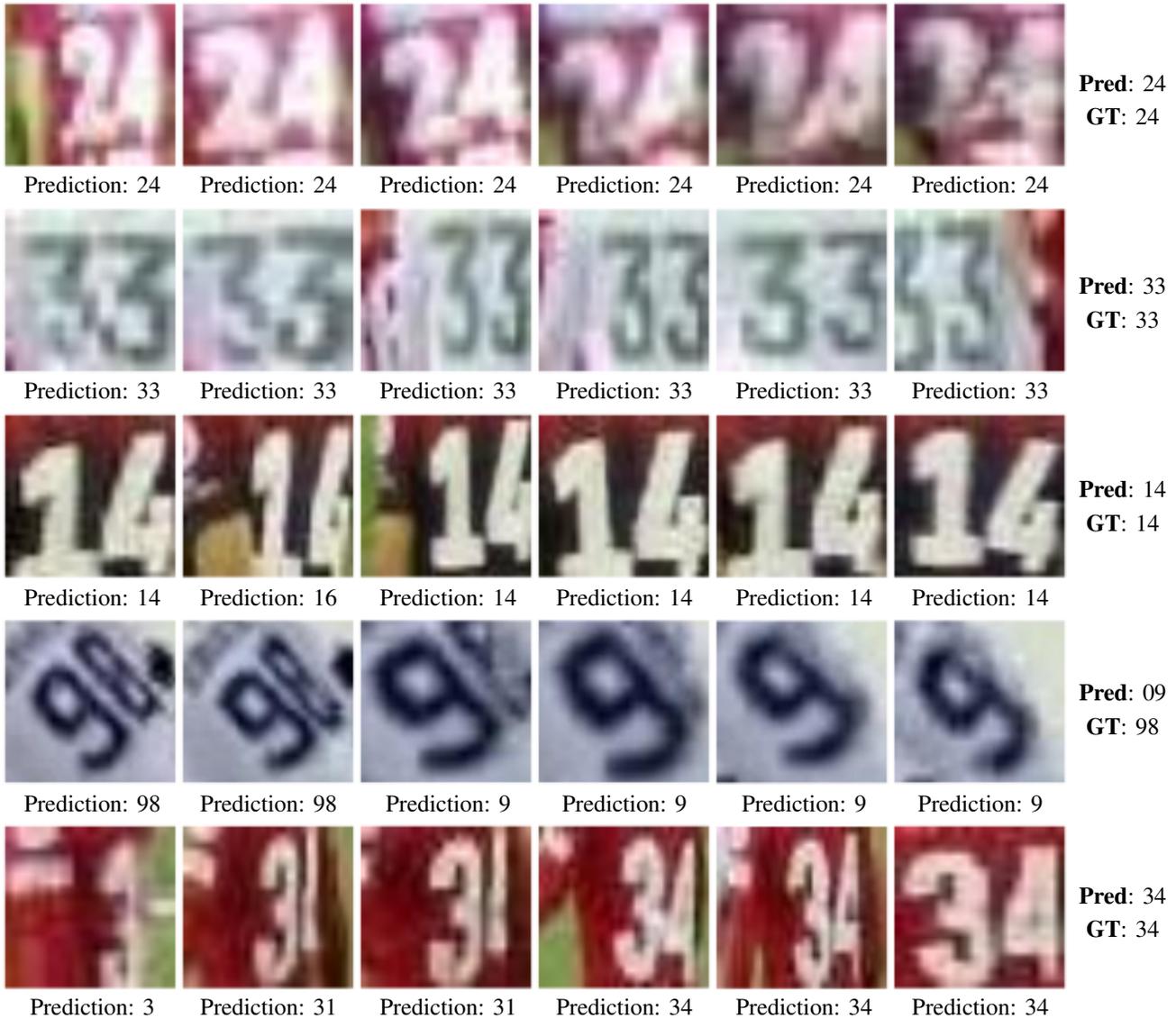

\textbf{Qualitative Results.} We present the qualitative results of our model on different tracklets from the datasets in Fig. \ref{fig:qual}. To obtain predictions for each image individually, we pass the same image feature $S$ times to our transformer decoder, where $S$ represents the number of tokens. For the overall tracklet prediction (TP), we pass the features of all the images within a tracklet to the transformer decoder. The results demonstrate that our model consistently predicts jersey numbers reliably, even in extremely blurry scenarios. 

\textbf{Spatial Backbone.} To understand the effectiveness of MAEs and the domain-guided masking strategy, we conduct an experiment with different backbones and masking strategies on the SoccerNet dataset as shown in Table \ref{tab:heads}. The results illustrate that leveraging MAEs as spatial feature extractors results in a 12.2\% increase compared to conventional backbones. This performance improvement is mainly credited to the self-supervised objective used in its pre-training stage. Furthermore, the observed accuracy boost of 1.84\% with the use of motion blur as the masking strategy validates our hypothesis that a domain-guided masking strategy enhances robustness to real-world broadcast data. 

\subsection{Ablation Study} \label{ablation}

\textbf{Keyframe Identification Module.} We evaluate the performance of our model on all 3 datasets with and without the processing from the KfID module in Table \ref{tab:preprocess}. Additionally, we quantitatively demonstrate the impact of our modifications in the JNL and GHC components on the overall improvement of the model performance in Table \ref{tab:KfID}.

\begin{table}[t]
  \centering
  \caption{Results with and without KfID Module. ($\dagger$) - with the KfID module}
  \label{tab:preprocess}
  \begin{tabular}{lll}
    \toprule
    \textbf{Dataset} & \textbf{Test Acc} & \textbf{Challenge Acc}\\
    \midrule
    Ice Hockey & 61.71 & -  \\
    Baseball & 88.43 & - \\
     \textbf{SoccerNet} & \textbf{35.65} & \textbf{35.98}  \\
    \midrule
    Ice Hockey ($\dagger$) & 96.79 $\textcolor{green}{\mathbf{{\uparrow}35.08}}$ & -\\
    Baseball ($\dagger$) & 94.70 $\textcolor{green}{\mathbf{{\uparrow}5.73}}$ & - \\
    \textbf{SoccerNet} ($\dagger$) & \textbf{77.31} $\textcolor{green}{\mathbf{{\uparrow}41.66}}$ & \textbf{81.92} $\textcolor{green}{\mathbf{{\uparrow}45.94}}$ \\
    
    \bottomrule
  \end{tabular}
\end{table}

The results in Table \ref{tab:preprocess} underscore the importance of the KfID module, as its incorporation leads to an increase of 35.08\%, 41.66\% and 5.73\% in the test accuracies of ice hockey, SoccerNet and Baseball datasets respectively. 
This shows the amount of spurious frames in real-world data and necessitates the need to identify useful detections.  

Table \ref{tab:KfID} illustrates the efficacy of our modified KfID as it improves our overall model accuracy by 1.84\%. Further examining the findings presented in Table \ref{tab:KfID} provides valuable insights into the significance of each implemented design change within the KfID module. Notably, the integration of our GHC module yields a 1.39\% enhancement in overall accuracy, underscoring the limitations associated with relying solely on color for clustering frames and emphasizing the need for deep features. Additionally, the incorporation of our JNL module demonstrates a 0.77\% improvement over the original JNL module in accuracy on the SoccerNet dataset. This underscores the critical role of robustly detecting smaller digits, thereby contributing to the refinement of our model's performance.

\begin{table}[t]
\caption{Effect of different proposed components of the KfID module on the overall accuracy in the SoccerNet dataset}
\label{tab:KfID}
\centering
\begin{tabular}{@{}lcc@{}}
\toprule
\textbf{Module} & \textbf{GHC old} & \textbf{GHC new(ours)}\\
\midrule
     JNL old & 75.47 & 76.54\\
     JNL new(ours) & 75.92 & \textbf{77.31}\\
\bottomrule
\end{tabular}
\end{table}

\textbf{Ablations on Siamese Loss.} 
We explore and evaluate different feature extractors for the Siamese loss $\mathcal{L}_{\textrm{siamese}}$ to assess their influence on the loss function and determine the optimal feature extractor. Furthermore, we compare different similarity and distance metrics to determine the most effective distance metric, which leads to quicker convergence and superior results. These experiments were conducted on the SoccerNet dataset, and the results are depicted in Table \ref{tab:ploss}.

\begin{table}[t]
\caption{Impact of feature extractors and metrics for $\mathcal{L}_{\textrm{siamese}}$ on our overall model performance}
\label{tab:ploss}
\centering
\begin{tabular}{@{}l|ccc@{}}
\toprule
\textbf{Feature Extractor} & $\ell_2$-loss & $\ell_1$-loss & Cosine Similarity\\
\midrule
     VGG & \textbf{76.30} & 76.21 & 74.52\\
     ResNet & 76.45 & \textbf{77.31} & 74.90\\
     InceptionNet & 75.84 & \textbf{75.93} & 74.66\\
     AlexNet & 74.38 & \textbf{74.41} & 73.93\\   
\bottomrule
\end{tabular}
\end{table}


Table \ref{tab:ploss} illustrates the efficacy of $\ell_1$-loss over other similarity metrics such as the cosine similarity and $\ell_2$ norm. The lower performance of cosine similarity can be attributed to its focus on the angle between two vectors rather than their magnitudes. This characteristic is less advantageous for image reconstruction tasks, where the magnitude of each feature holds more importance than the angle between them. Among the other two metrics, we hypothesise that the superior performance of $\ell_1$-loss could be attributed to the sparsity induced by this norm. This ensures that prominent features such as the edges in images are captured effectively. Additionally, the robustness of $\ell_1$-loss to outliers could contribute to its success. The table also demonstrate using ResNet for the Siamese objective yields superior results.

\section{Conclusion} \label{conclusion}

In this work, we address critical questions concerning the masking policy of MAEs for robust feature extraction in the realm of jersey number recognition. By introducing a novel domain-guided masking strategy, we devise a spatiotemporal network utilizing MAEs and temporal transformers to counter the issues of motion blur and occlusion for the player identification task. Additionally, we refine the existing KfID module to extract more reliable keyframes from the tracklet data. To tackle the issue of limited labelled data resulting due to the KfID module, we leverage a unique keyframe fusion approach to further augment the data. Through quantitative evaluation on different sports including Soccer, Ice Hockey and Baseball, we demonstrate the superior performance of our proposed approach against existing methods. Meticulous ablation studies on the impact of the KfID module and different masking strategies validate the state-of-the-art performance of our network on jersey number recognition.

Future research works involve experimenting with various masking strategies tailored to specific domains, thereby validating the significance of domain-guided masking approaches for robust feature extraction in MAEs.

\section{Acknowledgement}
We acknowledge the support of Stathletes and the Baltimore Orioles through the Mitacs Accelerate Program, as well as the Natural Sciences and Engineering Research Council of Canada (NSERC). Additionally, we express our appreciation to the Digital Research Alliance of Canada for their hardware support.


\bibliographystyle{IEEEtran}
\bibliography{main.bib}

\end{document}